\documentclass[10pt]{igs}

\usepackage{natbib}

\usepackage{amsmath,amstext,amsopn,amssymb,amsbsy}

\usepackage{tabularray}
\usepackage{multirow}
\usepackage{graphicx}
\usepackage{caption}
\usepackage{subcaption}
\usepackage{comment}
\usepackage{natbib}
\usepackage{xcolor}
\usepackage{graphics}
\usepackage{graphicx}
\usepackage{hhline}
\usepackage{bigstrut}
\usepackage{graphicx,subcaption,ragged2e}
\usepackage{colortbl}
\usepackage{float}
\usepackage{subcaption}
\usepackage{outlines}
\usepackage{adjustbox}

\begin{document}

\begin{frontmatter}


\title{Assessment of Autism and ADHD: A Comparative Analysis of Drawing Velocity Profiles and the NEPSY Test}


\author{Sol FORTEA\thanksref{label1}}
\author{, Alejandro GARCIA-SOSA\thanksref{label2}}
\author{, Paula MORALES-ALMEIDA\thanksref{label1}}
\author{and Cristina CARMONA-DUARTE\thanksref{label2}}
\address[label1]{Departamento de Psicología, Sociología y Trabajo Social} 
\address[label2]{Instituto para el Desarrollo Tecnológico y la Innovación en Comunicaciones IDeTIC\\ 
Universidad de Las Palmas de Gran Canaria, Spain\\
\{sol.fortea, paula.morales, alejandro.sosa, cristina.carmona\}@ulpgc.es\\}


\end{frontmatter}

%
%

%
%

%


%
%


%

%



\begin{abstract}
The increasing prevalence of Autism Spectrum Disorder and Attention-Deficit/ Hyperactivity Disorder among students highlights the need to improve evaluation and diagnostic techniques, as well as effective tools to mitigate the negative consequences associated with these disorders. With the widespread use of touchscreen mobile devices, there is an opportunity to gather comprehensive data beyond visual cues. These devices enable the collection and visualization of information on velocity profiles and the time taken to complete drawing and handwriting tasks. These data can be leveraged to develop new neuropsychological tests based on the velocity profile that assists in distinguishing between challenging cases of ASD and ADHD that are difficult to differentiate in clinical practice. In this paper, we present a proof of concept that compares and combines the results obtained from standardized tasks in the NEPSY-II assessment with a proposed observational scale based on the visual analysis of the velocity profile collected using digital tablets.



\end{abstract}


\section{Introduction}

Neurodevelopmental disorders have a significant impact on individuals, affecting various aspects of their development, including personal, social, academic, and occupational domains, according to DSM-5 (\cite{DSM-5}). 
Differentiating between Autism Spectrum Disorde (ASD) and Attention Deficit Hyperactivity Disorder (ADHD) can be challenging due to overlapping symptoms. However, research suggests that there are enough differences to justify separated diagnostic categories (\cite{Antshel}). Children with  ASD face difficulties in social communication, exhibit repetitive and restricted behaviours and often encounter learning challenges and difficulties with executive and motor coordination during their school years. Similarly, individuals with ADHD experience deficits in attention and impulse control, which also impact executive functioning across cognitive, emotional, and motor domains within the school setting. 
In some cases, considering a variety of motor parameters may be useful for the differential diagnosis.

In the field of autism, 
there is increasing recognition among authors of the need to consider motor difficulties. Recent studies have shown that motor impairments are prevalent in individuals with ASD. For example, \cite{Zampella2021},
found that up to 87 per cent of individuals with ASD exhibit clinically significant and widespread motor difficulties. Similarly, \cite{Gandotra}
concluded that deficits in basic motor skills can serve as an early marker of ASD, highlighting the importance of further studies in this area. The Lancet Commission's recommendations for the future of autism research and clinical practice include the assessment of motor skills (\cite{Lord}).
However, motor impairments have not been included in the recognized diagnostic criteria by the international scientific community (APA, 2013), nor have they been dismissed like the symptom referring to families of high intelligence. Most research on motor impairments in autism has focused on basic motor skills. 

To effectively address the unique needs of individuals with neurodevelopmental disorders, comprehensive and objective assessment tools are essential. One widely used test in this context is the NEPSY-II (\cite{Nepsy-II}). The Nepsy evaluates attention and executive functions, visuospatial functions, language and communication, sensorimotor functions, memory and learning, and social perception.
The child is presented with two-dimensional figures and is tasked with reproducing them on paper using a pencil. To successfully complete this task, the child must perceive and analyze the visuospatial relationships of the figures and convert this mental representation into action. The child's visuospatial analysis can be either global, capturing the overall shape of the figure, or local, focusing on the details but not perceiving the visual form of the figure, or a combination of both. Visual-motor control plays a crucial role in accurately copying the figures.
To assess a figure, the evaluator has to assign one point if the subject's execution follows the fixed rules by the test, or zero otherwise.

By employing such observation tests, professionals can design intervention programs tailored to children with alterations in assessed domains. However, in order to develop specific strategies and interventions, it is necessary to complement the NEPSY-II evaluation with additional tools that delve deeper into the execution process and identify the neuromotor signals involved in the writing processes.

On the other hand, the increasing popularity of mobile devices equipped with touch screens provides new opportunities for collecting comprehensive data that surpass traditional visual cues. These devices enable the collection of information on the velocity and time taken to complete drawing and handwriting tasks. By leveraging the capabilities of digital tablets, valuable insights could be gained into underlying motor impairments that may not be easily detected through conventional assessments (\cite{Silva, Hudry}). While graphomotor skills, such as writing legibility and velocity, have been frequently investigated (\cite{vanDeBos, rejeanAdhd}), the role of movement parameters, including velocity, in the detection and differential diagnosis of autism compared to other neurodevelopmental disorders remains understudied.

This paper presents a pilot study that aims to compare and complement the results obtained from standardized tasks in the NEPSY-II assessment with an observational scale based on a set of proposed parameters for assessing velocity profiles. The primary goal is to gather additional information on how children learn and execute patterns by analyzing the shape and velocity profiles obtained from handwriting or drawing tasks conducted on a digital tablet screen. We hypothesize that children with ASD are more likely to focus on copying the figure rather than simplifying the pattern, while children with ADHD may experience challenges in coordinating their movements and accurately controlling the shape they intend to draw.

The paper is structured as follows: Section \ref{sec:method} explains the methodology employed in this study. Section \ref{sec:exp} presents the experimental results and discussion and finally, in Section \ref{sec:conc}, we draw conclusions and discuss avenues for future research.

\section{Method}
\label{sec:method}
\subsection{Participants}


This pilot study included a total of twelve participants, with six having neurodevelopmental disorders (five with ASD and one with ADHD), and six participants with typical development (HC) as a comparison group (see Table \ref{tab:subjects}). 

Out of the six subjects with neurodevelopmental disorders (ND), five have been diagnosed with ASD, and only one has been diagnosed with ADHD. All subjects except Subject 4 have the ability to read and write, with Subject 4 being in the early stages of learning.
The autism spectrum is characterized by its broad range, which is reflected in the subjects of this sample. Cases 1 and 2 have been diagnosed with autism without cognitive deficits or ADHD. Case 3 exhibits a mild cognitive delay. Case 5 demonstrates high intellectual capacity (high IQ), while Case 6 has average intelligence (average IQ) and also presents comorbidity with ADHD. All of the subjects underwent evaluations conducted by clinical experts proficient in the administration of Autism Diagnostic Observation Schedule, version 2 (ADOS-2) (\cite{ADOS}), Autism Diagnostic Interview Revised (ADI-R) (\cite{adir}), and Wechsler intelligence scale for children-fifth edition (WISC-V) (\cite{wisc-V}) assessments. ADOS allows us to determine a specific level of severity, which, in subjects 1,2,3, 5 and 6, is the lowest level (level 1) in all cases, and 3 falls into level 2, likely due to her below-average IQ. 

The children with neurodevelopmental disorders are currently receiving psychoeducational treatment at a Child Psychology Health Center, which indicates their active involvement in interventions and support services. Furthermore, these children attend primary schools and reside with their families, reflecting their typical living and educational environments. The data were collected following ethical approval. Parents of the participants signed written informed consent forms before participation.

\begin{table}[ht!]
  \centering
 
  \caption{Subjects number by diagnosis and age} 
  \begin{adjustbox}{width=0.5\linewidth,center}
    \begin{tabular}{|l|c|c|c|l|c|c|}
    \hline
     \multicolumn{1}{|c|}{\textbf{Subject}} & \multicolumn{1}{c|}{\textbf{Diagnostic}} & \multicolumn{1}{c|}{\textbf{Age}} & & \multicolumn{1}{|c|}{\textbf{Subject}} & \multicolumn{1}{c|}{\textbf{Diagnostic}} & \multicolumn{1}{c|}{\textbf{Age}} \bigstrut\\
    \hline
   1   & ASD & 13 & & 7   & Control & 7 \bigstrut \\
    \hline
      2   & ASD & 12 & & 8  &   Control & 7  \bigstrut\\ 
    \hline
       3   & ASD with low IQ & 11& &  9   & Control & 7  \bigstrut\\
    \hline
       4   & ADHD & 7& & 10  &  Control & 8 \bigstrut\\
    \hline
      5   & ASD with high IQ  & 8 & & 11   & Control & 9 \bigstrut\\
    \hline
     6   & ASD with ADHD& 11 &  & 12   & Control & 7 \bigstrut\\
    \hline
    \end{tabular}%
   
  \label{tab:subjects}%
 \end{adjustbox}
\end{table}%

\subsection{Data recording}
An app was developed for an iPad Pro (5th generation) with a screen size of 12.9 inches, which was used to display the different tasks to the children and record their handwriting. The children were instructed to use an Apple Pencil to copy the figures displayed on the screen.

\subsection{Tasks}

In this pilot study, we selected three specific tasks to visually assess the motor characteristics of the participants. In this study, from the Nepsy test, we have considered only one task from the visuospatial function's domain, specifically the design copy task. From this task, only the first two elements are currently feasible for implementation: the copy of the circle and the copy of the square. These elements can be drawn without lifting the stylus from the Tablet, which is a requirement for data collection since the iPad can only record the movement on the tablet screen, and they are the simplest ones in the Nepsy test and are relatively straightforward for individuals who already know how to write. They provide valuable information regarding the acquisition or lack thereof of a motor pattern. The third task was selected based on its use in other studies related to neurodegenerative diseases, as it involves pattern repetition and aligns with our hypothesis. Thus, the three selected tasks were as follows: Task 1: Copying a circle, Task 2: Copying a square, and Task 3: Writing "elelelel".

\subsection{Criteria of task 1 and 2 based on standardized measures: Nepsy II}
\label{subsec:nepsy_scale}
NEPSY-II is a clinical tool that has been standardized for face-to-face administration. To ensure an objective assessment, the test provides clear guidelines on how to evaluate each task. 
In this study, serving as a proof of concept, we specifically focus on assessing three tasks as defined in the previous subsection. 
For this pilot study, items 3 and 4 of the subtest Design Copying (DC), namely the circle copy and square copy, respectively, are selected (task 1, task 2). In the circle copy task, the evaluation focuses on whether the drawing represents a circular shape, the presence or absence of interruptions in the figure, the presence of straight lines within the circle, and the overall proportionality. In the square copy task, the evaluation assesses whether the drawing represents a geometric figure with four connected sides, its alignment with the horizontal plane, and the absence of extensions or additional elements on the sides. 

These evaluations are based on the parameters provided by the NEPSY-II. The maximum scores obtained will be 5 points for the copy of the circle and 7 points for the copy of the square, depending on whether or not they meet a series of criteria (1 point or 0 points each) provided by the NEPSY-II.

\subsubsection{Criteria of task 3}

Task 3 involves a simple copy of the series "elelelel" in calligraphy. Unlike the standardized tasks of the NEPSY-II, this task is not standardized. However, we evaluated them using a similar criteria to those used in tasks 1 and 2. We awarded 1 point for each criterion achieved, with a maximum score of 5 points:
\begin{itemize}
 \item \textbf{General Copy}: We appreciate that the task involves two letters, 'e' and 'l', chained together in calligraphy.


\item \textbf{Processing Copy:}
\begin{itemize}
\item Motor A. The letters have a change of direction at the top.
\item Motor B. There is no section of the letters that is a horizontal line longer than the height of the tallest letter.
\item Global C. The letters 'e' and l' alternate repeatedly, at least three times.
\item Global D. The sequence of letters is located in an area of 30º with respect to the horizontal axis.

\end{itemize}
\end{itemize}

\subsection{Velocity observation scale: proposed procedure}
\label{sec:Par}

To further understand how patterns are learned and executed, we propose to observe manually certain parameters in the velocity signal representation, in addition to using the Nepsy scale, to quantify the correctness of the trace execution. In healthy adults, we can observe that movements are characterized by peaks within the velocity profile (\cite{PlamondonII,Plamondon1,PlamondonIII,TemporalEvolution}),  and each change of direction is accompanied by a minimum in velocity. Additionally, individual letters or groups of letters are executed together, with greater pauses occurring between groups.
Based on these observations, we propose four parameters to evaluate movement performance, scored similarly to how it is done in the Nepsy test:

\begin{itemize}

\item Velocity pattern repetitions (VPR): This parameter assesses whether there is a cyclic pattern that repeats in the velocity profile.If a pattern is observed, 1 point is awarded; otherwise, 0 points are given. 

\item Velocity changes (VC): This parameter evaluates whether the velocity remains constant or fluctuates throughout the task. It is observed when the peak velocities reached are similar in height.If the velocity peaks height are similar 1 point is awarded; otherwise, 0 points are given.

\item Pauses (P): This parameter examines if the duration of the pauses are similar. If duration in the stops or minimun of velocity are similar, 1 point is awarded; otherwise, 0 points are given. In the case of the circle, 0 points are awarded if clear peaks with different distances between them are observed. 

\item Time (T): This parameter measures the time taken to complete the task. 4 seconds or less for the circle and the square, or 8 seconds or less in the repetitions of three "el" scores 1 point; otherwise, 0 points are given.

\end{itemize}

Therefore, the maximum score for each subject in each parameter will be 4 points, with a maximum possible score of 12 points. 


\subsection{Stastitical Analysis}
\label{sec:stastitical}
To determine the discriminative utility of the combination of two velocity parameters and identify the potential number of possible clusters, we employed the k-means clustering algorithm (\cite{k-means}) and Silhouette (\cite{silhouettes}). Silhouette values are represented as a vector of size n-by-1, where n represents the number of points in the dataset. These values range between -1 and 1. A silhouette value measures the similarity of a point to other points within its own cluster compared to points in other clusters. A high silhouette value indicates that a point is well-matched to its own cluster and poorly matched to other clusters. The range of -1 to 1 allows for the evaluation of how well-defined and distinct the clusters are, with positive values indicating strong cluster membership and negative values suggesting possible misclassification. 

\section{Experimental results}

\label{sec:exp}

\subsection{Data}
\begin{figure}[h!] 
\captionsetup[figure]{justification=Centering}
\centering
\begin{subfigure}[t]{0.45\textwidth}
    \includegraphics[width=\linewidth]{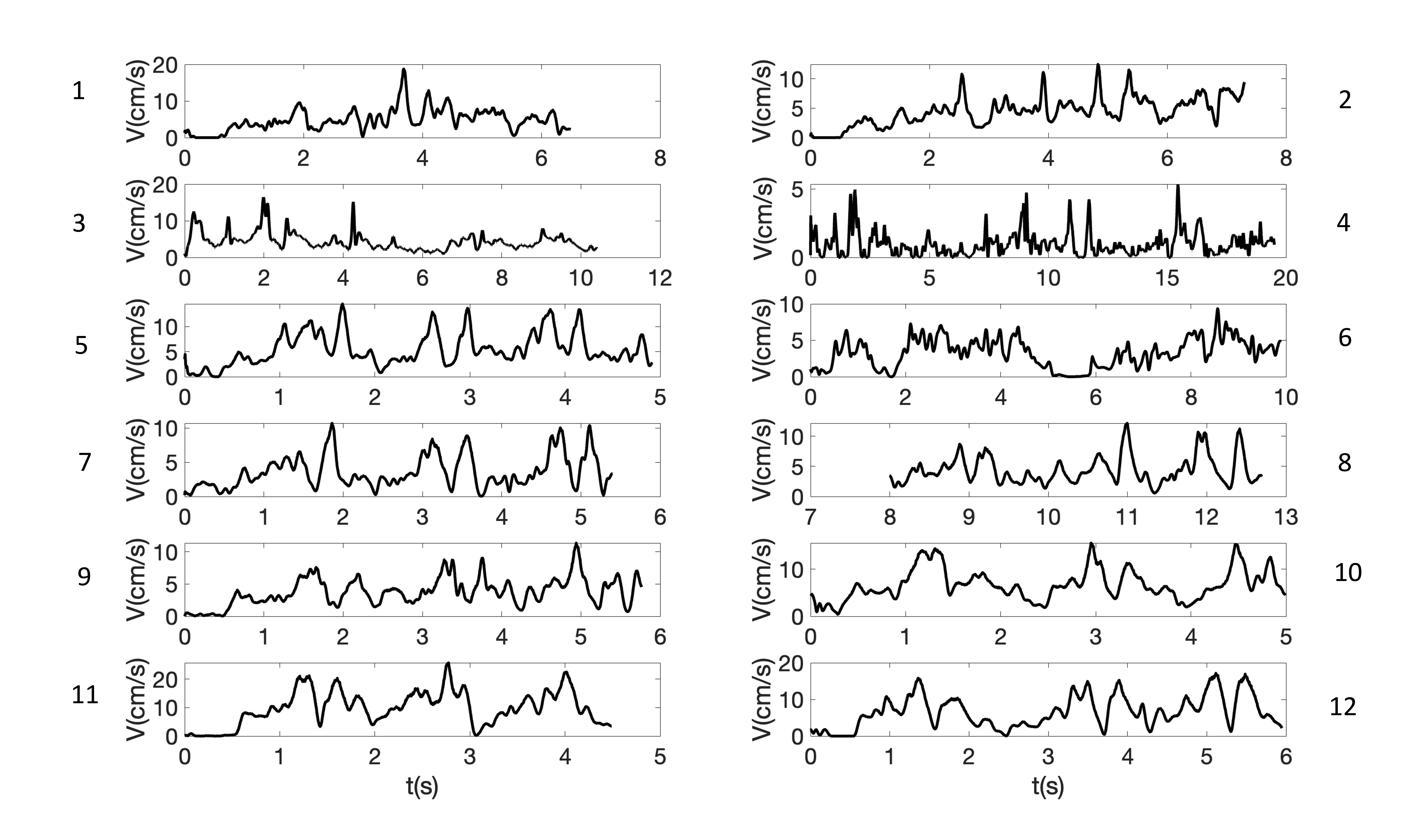}  
    \caption{Task 3. Velocity.}
\end{subfigure}
\begin{subfigure}[t]{0.45\textwidth}
    \includegraphics[width=\linewidth]{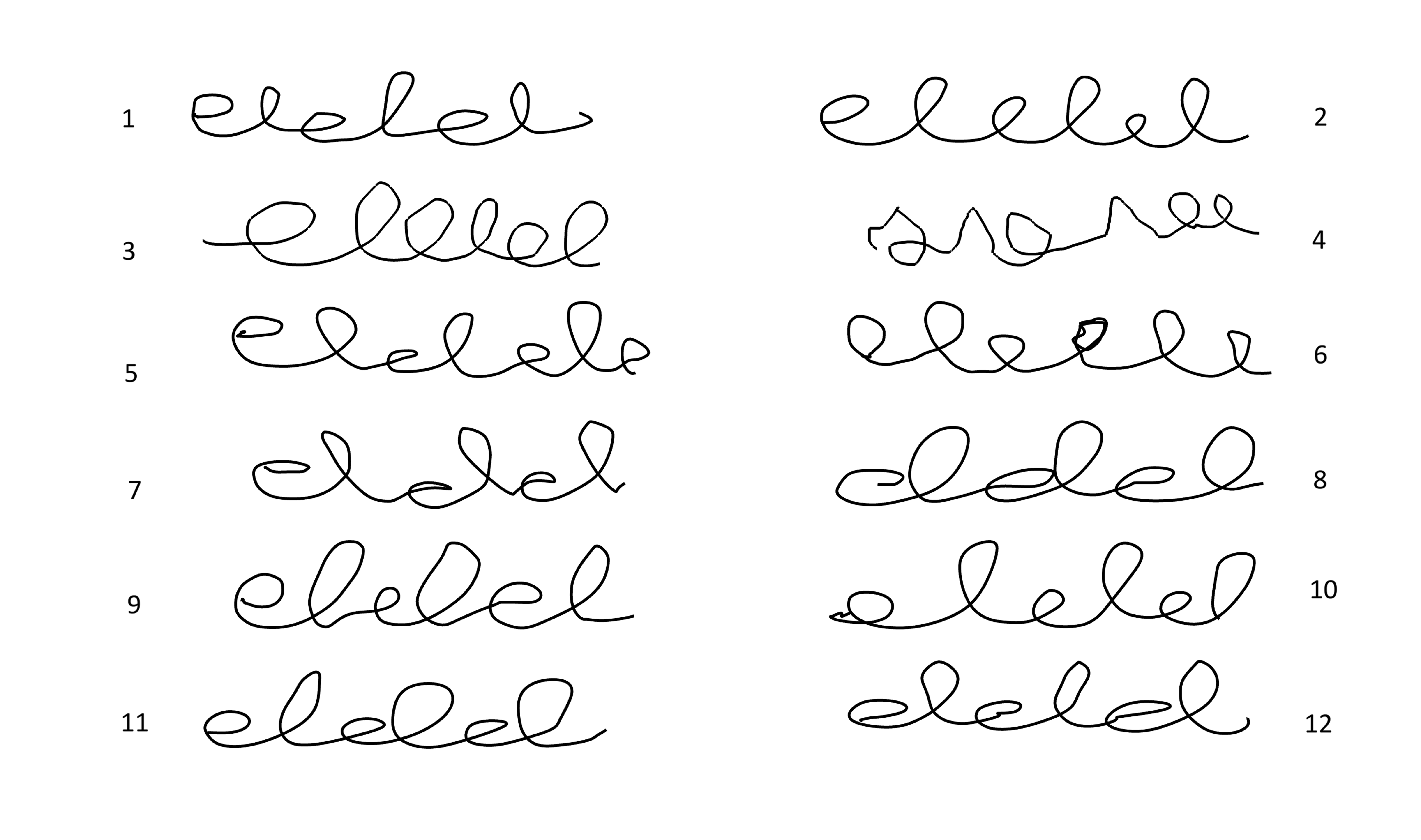}
    \caption{Task3. Draw.}
\end{subfigure}
\caption{Task 3. Zoom to only 3 repeticions }

\label{Zoom}
\end{figure}

In Figure \ref{Zoom},
we can observe a sample of the data captured from the 12 subjects described in the previous section.

\subsection{Nepsy observation scores}


In the NEPSY scale, the shapes of the drawings were evaluated using the  NEPSY scale shown in Subsection \ref{subsec:nepsy_scale}, to obtain the results shown in Table \ref{NepsyTable}.

We can observe in Table \ref{NepsyTable} that most subjects achieved maximum scores, except for subjects 3, 4, and 5. Subject 4 shows the greatest difference compared to the control group (HC), which could be because he has not yet acquired complete proficiency in handwriting performance.

\begin{table}[h!]
\renewcommand\arraystretch{1.2}
\caption{Observation scale results}
\centering

\begin{subtable}[h]{0.45\textwidth}
\caption{Nepsy results }
\label{NepsyTable}
\resizebox{7cm}{!} {
\begin{tabular}{| c | c | c | c | c | c | c | c | c | c | c | c | c | c|}
\hline 
 \multicolumn{2}{|c|}{}&  \multicolumn{6}{|c|}{ND} &\multicolumn{6}{|c|}{HC} \\
\hline Task  &Subject                & 1 & 2 & 3 & 4 & 5 & 6 & 7 & 8 & 9 & 10 & 11 & 12 \\
\hline
\multirow{6}{*}{Circle} 
& Copia General		    & 1 & 1 & 1 & 1 & 1 & 0 & 1 & 1 & 1 & 1 & 1 & 1 \\
& Motor A               & 1 & 1 & 1 & 1 & 1 & 1 & 1 & 1 & 1 & 1 & 1 & 1 \\
& Motor B               & 1 & 1 & 1 & 1 & 1 & 1 & 1 & 1 & 1 & 1 & 1 & 1 \\
& Global C              & 1 & 1 & 1 & 1 & 1 & 0 & 1 & 1 & 1 & 1 & 1 & 1  \\
& Global D              & 1 & 1 & 1 & 1 & 1 & 1 & 1 & 1 & 1 & 1 & 1 & 1 \\
\cline{3-14}
& \textbf{Total}        & 5 & 5 & 5 & 5 & 5 & 3 & 5 & 5 & 5 & 5 & 5 & 5 \\
\hline
\multirow{8}{*}{Square}
&Copia General		 & 1 & 1 & 1 & 1 & 0 & 1 & 1 & 1 & 1 & 1 & 1 & 1 \\

& Motor A            & 1 & 1 & 1 & 1 & 1 & 1 & 1 & 1 & 1 & 1 & 1 & 1 \\
& Motor B            & 1 & 1 & 1 & 1 & 1 & 1 & 1 & 1 & 1 & 1 & 1 & 1 \\
& Global C           & 1 & 1 & 1 & 1 & 1 & 1 & 1 & 1 & 1 & 1 & 1 & 1 \\
& Global D           & 1 & 1 & 1 & 1 & 1 & 1 & 1 & 1 & 1 & 1 & 1 & 1 \\
& Local E            & 1 & 1 & 1 & 1 & 1 & 1 & 1 & 1 & 1 & 1 & 1 & 1 \\
& Local F            & 1 & 1 & 1 & 1 & 1 & 1 & 1 & 1 & 1 & 1 & 1 & 1 \\
\cline{3-14}
&  \textbf{Total}    & 7 & 7 & 7 & 7 & 6 & 7 & 7 & 7 & 7 & 7 & 7 & 7 \\

\hline
\multirow{7}{*}{elelele}
& Copia General		    & 1 & 1 & 1 & 0 & 1 & 1 &  1 & 1 & 1 & 1 & 1 & 1 \\

& Motor A               & 1 & 1 & 1 & 0 & 1 & 1 & 1 & 1 & 1 & 1 & 1 & 1 \\
& Motor B               & 1 & 1 & 1 & 0 & 1 & 1 & 1 & 1 & 1 & 1 & 1 & 1 \\
& Global C              & 1 & 1 & 1 & 0 & 1 & 1 & 1 & 1 & 1 & 1 & 1 & 1 \\
& Global D              & 1 & 1 & 1 & 0 & 1 & 1 & 1 & 1 & 1 & 1 & 1 & 1 \\
\cline{3-14}
&  \textbf{Total}       & 5 & 5 & 5 & 0 & 5 & 5 & 5 & 5 & 5 & 5 & 5 & 5 \\
\cline{1-14}
 &  \textbf{Total}       & 17 & 17 & 17 & \textbf{12} & \textbf{16} & \textbf{15}& 17 & 17 & 17 & 17 & 17 & 17\\
\cline{1-14}

\end{tabular}
}
 \hfill
\end{subtable}

   \begin{subtable}[h]{0.45\textwidth}
\caption{Velocity observation scale results}
\resizebox{7cm}{!} {
\begin{tabular}{| c | c | c | c | c | c | c | c | c | c | c | c | c | c |}
\hline 
 \multicolumn{2}{|c|}{}&  \multicolumn{6}{|c|}{ND} &\multicolumn{6}{|c|}{HC} \\
\hline 

Task &Subject & 1 & 2 & 3 & 4 & 5 & 6& 7 & 8 & 9 & 10 & 11 & 12  \\
\hline
\multirow{4}{*}{Circle} & VPR		 & 0 & 0 & 1 & 1 & 1 & 0  & 1 & 1 & 1 & 1 & 1 & 1\\
& VC & 1 & 1 & 0 & 0 & 1 & 1   & 1 & 1 & 1 & 1 & 1 & 1\\
& P & 1 & 1 & 1 & 0 & 0 & 0   & 1 & 1 & 1 & 1 & 1 & 1\\
& T & 1 & 1 & 1 & 0 & 1 & 1    & 1 & 1 & 1 & 1 & 1 & 1\\
\cline{3-14}
& \textbf{Total} & 3 & 3 & 3 & 1 &3& 2  & 4 & 4 & 4 &4 &4&4 \\
\hline
\multirow{4}{*}{Square}& VPR		 & 0 & 0 & 1 & 0 & 1 & 1  & 1 & 1& 1 & 1 & 1 & 1\\
& VC & 0 & 1 & 1 & 0 & 0 & 1    & 1 & 1& 1 & 1 & 1 & 1 \\
& P & 1 & 1 &0 & 0 & 1 & 1      & 1 & 1 & 1 & 1 & 1 & 1\\
& T & 0 & 1 & 0 & 0 & 1 & 1      & 1 & 1 & 1 & 1 & 1 & 1 \\
\cline{3-14}
& \textbf{Total} & 1 & 3 & 2 & 0 & 3 & 3   & 4 & 4 & 4 & 4 & 4 & 4 \\

\hline
\multirow{4}{*}{elelele}& VPR		 
     & 0 & 0 & 0 & 0 & 1 & 0    & 1 & 1 & 1 & 1 &  1 & 1\\
& VC & 0 & 0 & 0 & 0 & 1 & 1    &1 & 1 & 1 & 1 & 1 & 1 \\
& P & 1 & 1 & 0 & 0 & 1 & 0   & 1 & 1 & 1 & 1 & 1 & 1\\
& T & 1 & 1& 0 & 0 & 1 & 0    & 1 & 1 & 1 & 1 & 1 & 1  \\
\cline{3-14}
& \textbf{Total} 
     &2& 2 & 0 & 0 & 4 & 1 &   4 & 4 & 4 & 4&4& 4\\
\hline
\multirow{4}{*}{Total Items}& VPR		
      & 0 & 0 & 2 & 1 & 3 & 1    & 3 & 3 & 3 & 3 &  3 & 3\\
& VC & 1 & 2 & 1 & 0 & 2 & 3    &3 & 3 & 3 & 3 & 3 & 3 \\
& P & 3 & 3 & 1 & 0 & 2 & 1   & 3 & 3 & 3 & 3 & 3 & 3\\
& T & 2 & 3 & 1 & 0 & 3 & 2    & 3 & 3 & 3 & 3 & 3 & 3 \\

\cline{2-14}

& \textbf{Total} & 6 & 8 & 5 & 1 & 10 & 7    & 12 & 12 & 12 & 12 & 12 & 12\\

\hline
\end{tabular}
}
\label{ResulVelo}
\end{subtable}
\end{table}

\subsection{Velocity observation scores}

Following the parameters proposed in Section \ref{sec:Par}, the results of the assessment of the velocity representation, as the sample shown in Figure  \ref{Zoom}, can be found in Table \ref{ResulVelo}.
In this case, the differences between the control group and the group with neurodevelopmental disorders (ND) are more pronounced. Subject 4 (ADHD) obtains the lowest score, as in Nepsy test. On the other hand, subject 5 exhibits a less significant difference, which will be further explained in the upcoming subsection.

\begin{figure}[h!] 

\centering

\begin{subfigure}[t]{0.32\textwidth}
   \includegraphics[width=\textwidth]{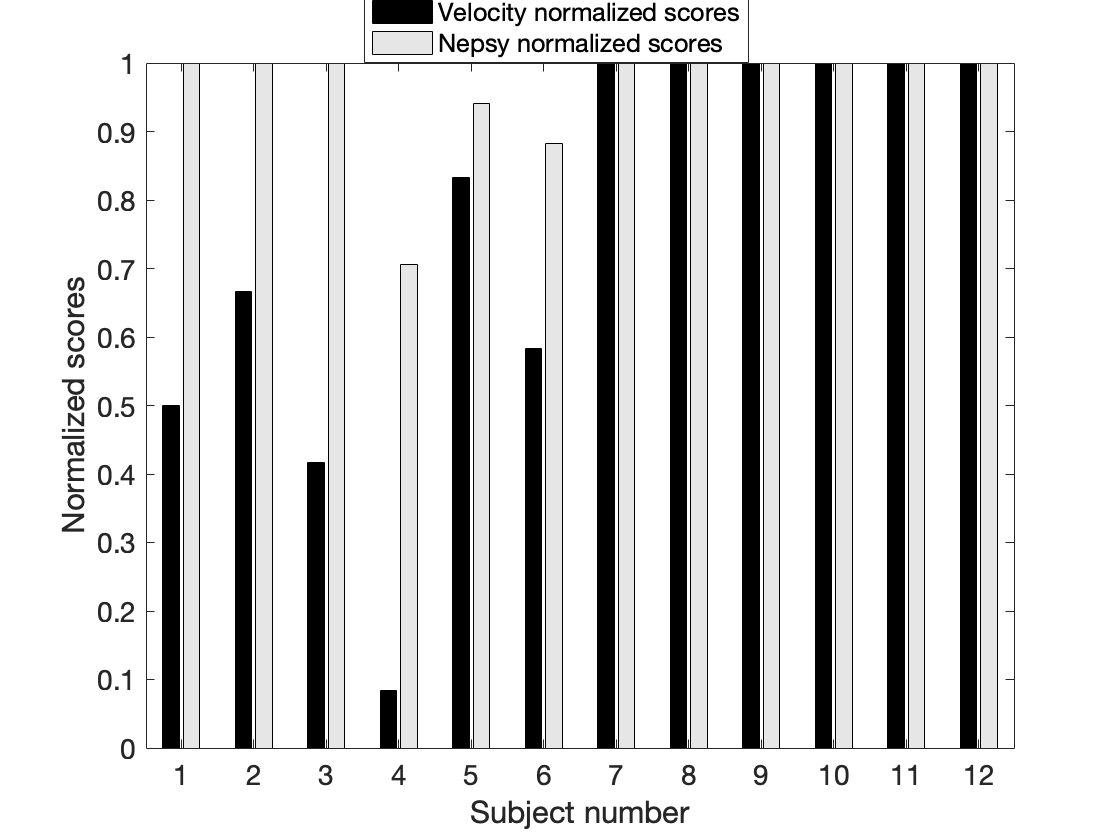}
    \caption{}
  \label{bars}
\end{subfigure}
\begin{subfigure}[t]{0.32\textwidth}
  \includegraphics[width=\textwidth]{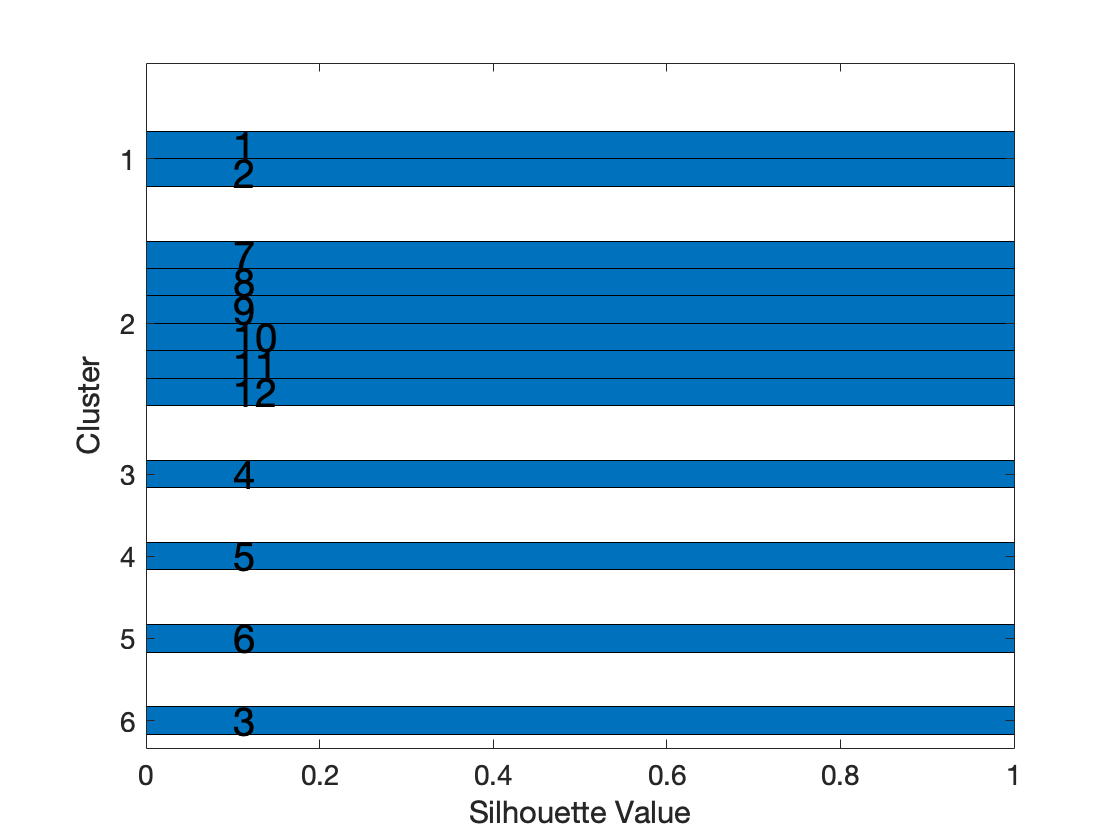}
    \caption{}
    \label{silhouette}
\end{subfigure}
\begin{subfigure}[t]{0.32\textwidth}
  \includegraphics[width=\textwidth]{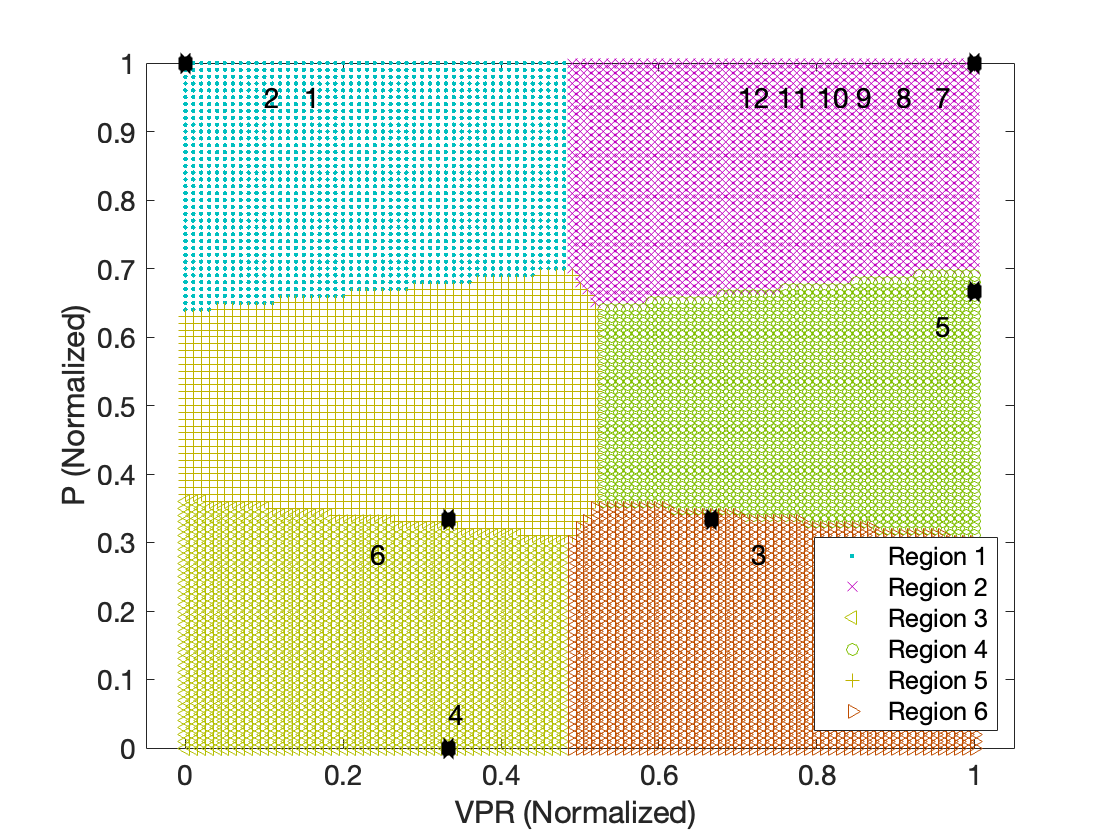}
    \caption{}
      \label{cluster}
\end{subfigure}
\caption{Clustering of the different subjects by VPR vs P: (a) Comparison of Nepsy and velocity results per each subject, (b) Silhouette values, (c) K-means result (VPR vs P) with 6 clusters. }

\end{figure}

If we use Kmeans and Silhouetter value (section \ref{sec:stastitical}) to analize the number of cluster convining VPR and P parameters, we can see the results in Figure  \ref{cluster} and \ref{silhouette}. We can observed the presence of six clusters with a significant difference, as the Silhouette value is one to all subjects (see Figure \ref{silhouette}), which means it is is well-matched to its own cluster and poorly matched to other clusters. 
The first cluster consists of individuals with ASD (Subjects 1 and 2), the second cluster includes all the children in the HC group, the third cluster represents individuals with ADHD, in the fifth one we find subject 5 with a high IQ and, and the last one, is subject 6 with ASD and low intelligence quotient.

\subsection{Discussion} 

The interpretation, following the NEPSY criteria as shown in Table \ref{NepsyTable} and Figure \ref{bars}, shows normal scores in all cases except for subjects 4, 5, and 6. These three subjects demonstrate difficulties in drawing the proposed shape, as it is shown in the Nepsy test scores. However, the velocity signals collected, as shown in Table \ref{ResulVelo} and Figure \ref{bars}, indicate neuromotor alterations in subjects 1 to 6. If we analyze Figure \ref{bars}, where the results of the NEPSY are compared with the velocity profile, it is observed that the control group shows the maximum scores in both tests. However, the differences in subjects with neurodevelopmental disorders are evident in the velocity profile test, as they do not demonstrate a consistent execution speed while performing the task. It is noteworthy that subject 4, with ADHD, shows the greatest differences, perhaps due to the alterations in executive functions characteristic of this population. Additionally, we can see that in the NEPSY test, the two children (subjects 4 and 6) diagnosed with ADHD present lower scores. This is an interesting matter that we must delve into further to confirm this suspicion.

The execution duration (T) (see Table \ref{ResulVelo}) is low in subjects 3 and 4. In the case of subjects 3 and 4, and 6 we can also observe an affected parameter P, which is related to the synchronization of the brain.

The VPR (Velocity Profile Ratio) is too low in subjects 1 and 2, with many peaks/movements without a pattern. This reflects that they have not internalized a motor pattern and are merely copying the drawing, rather than learning the task, as could be the case in individuals with ASD.

Subject 5, who has been diagnosed with ASD and high IQ, displays more similarities to the normal or ADHD subjects rather than those with ASD. This observation raises the possibility that his positive progress could be attributed to early intervention during early developmental stages and a typical educational path. Alternatively, it is also plausible that subject 5 may have another form of neurodevelopmental disorder with symptoms similar to ASD. Further investigation is needed to determine the exact nature of his condition and the factors contributing to their observed similarities.

These findings highlight the potential for obtaining additional information in assessments by observing these new parameters, thereby could prompt the development of novel standardized measures based on velocity observations.

\section{Conclusions and future work}
In this pilot study, a small sample of children was utilized to evaluate the effectiveness of a proposed test that combines the velocity profile analysis and selected Nepsy tests. The objective of the study was to explore the feasibility of using these tests to differentiate between different types of neurodevelopmental disorders. By examining how movements are organized and executed, the study aims to provide clinicians with more objective measures for assessment and diagnosis purposes.

As future work, the next step can be to apply recent advancements in computer tools, including machine learning and lognormal assessment, among others, to automate the calculation of scores for each assessment item proposed in this paper. By leveraging these tools, clinicians could obtain more accurate and objective assessments, leading to improved diagnosis and treatment for individuals with neurodevelopmental disorders.


\label{sec:conc}

\section{Acknowledgement}
This work has been supported by the Spanish project PID2021-122687OA-I00 / AEI  / 10.13039 / 501100011033 /FEDER, UE, and the  Canary Islands Employment Service, Investigo Programme  32/39/2022-0923131539, Recovery, Transformation and Resilience Plan - NextGeneration EU.
\bibliographystyle{apalike}
\bibliography{bib}


\end{document}